\documentclass[lettersize,journal]{IEEEtran}
\usepackage{amsmath,amsfonts}
\usepackage{array}
\usepackage{dcolumn}
\usepackage[caption=false,font=normalsize,labelfont=sf,textfont=sf]{subfig}
\usepackage{textcomp}
\usepackage{url}
\usepackage{verbatim}
\usepackage{graphicx}
\usepackage{cite}
\usepackage{framed,color}
\usepackage{balance}
\usepackage{booktabs}
\usepackage{pifont}
\newcommand{\tworowcell}[1]{\raisebox{-1.8ex}[0pt][0pt]{#1}}
\begin{document}

\title{Learning a Motion Manifold to Constrain Ambiguity in Unsupervised 4D Medical Image Interpolation}

\author{Haojin~Li, Huazhu~Fu, Heng~Li, and Jiang~Liu%
\thanks{This work was supported in part by the National Key R\&D Program of China under Grant 2024YFC2510800, the Shenzhen Medical Research Funds, China, under Grant D2602016, the National Natural Science Foundation of China under Grant 62401246, the Guangdong Basic and Applied Basic Research Foundation, China, under Grant 2025A1515010826, and the Shenzhen Science and Technology Program under Grants JCYJ20250604185805008 and JCYJ20240813095112017.}%
\thanks{Corresponding authors: Heng Li and Jiang Liu (e-mail: liheng@suat-sz.edu.cn; liuj@sustech.edu.cn).}%
\thanks{Haojin Li and Jiang Liu are with the Department of Computer Science and Engineering, Southern University of Science and Technology, Shenzhen 518055, China.}%
\thanks{Huazhu Fu is with the Institute of Advanced Intelligence and Computing, Agency for Science, Technology and Research (A*STAR), Singapore 138632.}%
\thanks{Heng Li is with Shenzhen University of Advanced Technology, Shenzhen 518107, China.}}

\markboth{IEEE Transactions on Medical Imaging}%
{Li et al.: Learning a Motion Manifold to Constrain Ambiguity}


\maketitle

\begin{abstract}

Unsupervised 4D medical image interpolation reconstructs intermediate anatomical states from temporally sparse observations without intermediate supervision. Distinct motion fields may attain comparable endpoint agreement while producing different intermediate anatomy, giving rise to motion ambiguity. This ambiguity is difficult to constrain when many independently adjustable motion variables must be estimated from the observed endpoints. We develop a framework that learns a motion manifold to restrict candidate motion fields while retaining sample-specific variation. Cross-Sample Spectral Basis Learning extracts recurring motion structure from endpoint-derived dense estimates and encodes it in a shared spectral basis. Sample-Conditioned Cayley Basis Reorientation uses predicted motion coordinates to rotate the effective motion directions for each endpoint pair without increasing the number of independent coordinates. Their composition defines a nonlinear family of complete velocity fields across endpoint pairs, accommodating anatomical and motion-state differences within a restricted representation. Integrating each predicted field supports bidirectional reconstruction at arbitrary intermediate times. Theoretical analysis establishes sufficient conditions under which endpoint observations distinguish nearby motions within the learned family. Experiments on ACDC and 4D-Lung demonstrate state-of-the-art performance among the evaluated methods trained without intermediate volumes and competitive results against methods using intermediate supervision.

\end{abstract}

\section{Introduction}

4D medical imaging captures anatomical change over time and supports functional assessment~\cite{meng2023deepmesh} and image-guided intervention~\cite{huttinga2021real}. Volumetric acquisitions, including dynamic CT and MRI, face trade-offs between temporal resolution, spatial fidelity, and modality-dependent constraints such as acquisition time and radiation exposure~\cite{reiter2025fourdimensional,demehri2023musculoskeletal}. Interpolating missing volumes offers a computational route to temporally denser anatomical observations without requiring equally dense acquisition~\cite{guo2020spatiotemporal,kim2024data}.

\begin{figure}[t]
\centering
\includegraphics[width=\columnwidth]{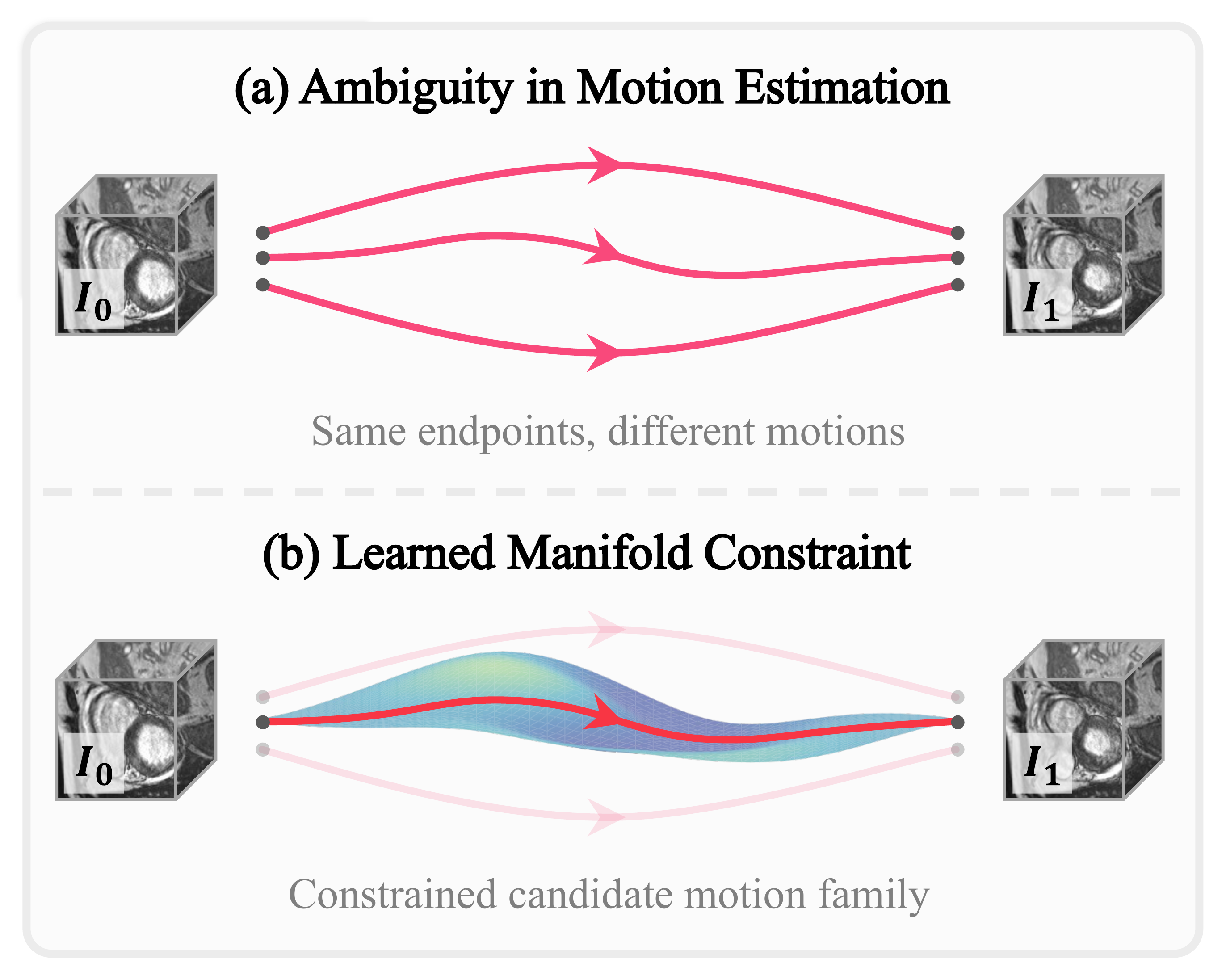}
\caption{Constraining motion ambiguity with a learned motion manifold. (a) The same endpoint volumes can admit different intervening motions. (b) Learning shared motion structure restricts the candidate motions available for endpoint fitting while accommodating sample-specific variation.}
\label{fig:motion_ambiguity}
\end{figure}

Unsupervised interpolation learns without intermediate-volume targets, deriving supervision from the observed endpoints~\cite{kim2024data}. Endpoint reconstruction evaluates anatomical alignment but does not specify the intervening motion. As shown in Fig.~\ref{fig:motion_ambiguity}, distinct motion fields may attain comparable endpoint agreement while producing different intermediate anatomy, giving rise to motion ambiguity~\cite{zhou2023exploring,zhong2026velocity}. Learned correspondence~\cite{balakrishnan2019voxelmorph,chen2022transmorph}, temporal consistency~\cite{kim2024data}, and generative priors~\cite{kim2022diffusion,you2025fb} guide estimation, yet endpoint fitting can retain many spatially distributed motion variables with only indirect evidence of how their changes should be coordinated.

Learning a motion manifold offers a representation-level route to constrain this ambiguity. It limits estimation to a family of complete motion fields described by a small set of motion coordinates~\cite{meila2024manifold}. These coordinates couple changes across the volume, allowing endpoint fitting to draw on recurring relationships between spatially separated regions. Statistical and learned models capture recurring anatomical motion~\cite{ehrhardt2011statistical,qiao2024cheart}, alongside variation across individuals~\cite{qiao2025personalized} and respiratory states~\cite{huang2024resolving}. A useful motion family should make this shared structure available as a constraint while accommodating the anatomy and motion state of each endpoint pair.

Two challenges arise in learning this family from endpoint observations. 1) Shared motion structure must be inferred without intermediate motion targets. Endpoint-derived dense estimates provide motion evidence, but each is only one possible explanation of an endpoint pair. Their recurring variation must be identified across samples to establish a shared reference. 2) Sample-specific motion must remain expressive within the learned family. The restriction should accommodate changes in anatomy and motion state without restoring a large number of independently adjustable motion variables.

We develop a framework that learns a motion manifold from endpoint observations and uses it to construct continuous deformations for interpolation. Cross-Sample Spectral Basis Learning summarizes endpoint-derived dense estimates in a shared spectral basis, providing a data-derived reference for motion synthesis. Sample-Conditioned Cayley Basis Reorientation uses motion coordinates predicted from each endpoint pair to rotate the effective directions within this representation. This coordinate-dependent reorientation supports nonlinear motion variation while keeping the number of independent coordinates fixed. The learned family constrains ambiguity by restricting the motion fields considered in endpoint fitting. Integrating the resulting velocity field supports bidirectional reconstruction at arbitrary query times. Theoretical analysis characterizes conditions under which endpoint observations distinguish nearby motion coordinates.

The contributions of this work are summarized as follows:

\begin{itemize}
    \item We develop a framework that learns a motion manifold to constrain ambiguity in unsupervised 4D medical image interpolation while retaining sample-specific motion variation.
    \item We introduce Cross-Sample Spectral Basis Learning, which extracts recurring motion structure from provisional endpoint-derived estimates to establish the shared basis of the motion manifold.
    \item We introduce Sample-Conditioned Cayley Basis Reorientation, which enables sample-specific motion expression by varying effective motion directions at a fixed number of independent motion coordinates.
    \item Experiments on ACDC and 4D-Lung demonstrate state-of-the-art accuracy among the evaluated methods trained without intermediate volumes and competitive performance against methods using intermediate supervision.
\end{itemize}

\begin{figure*}[!t]
\centering
\includegraphics[width=\textwidth]{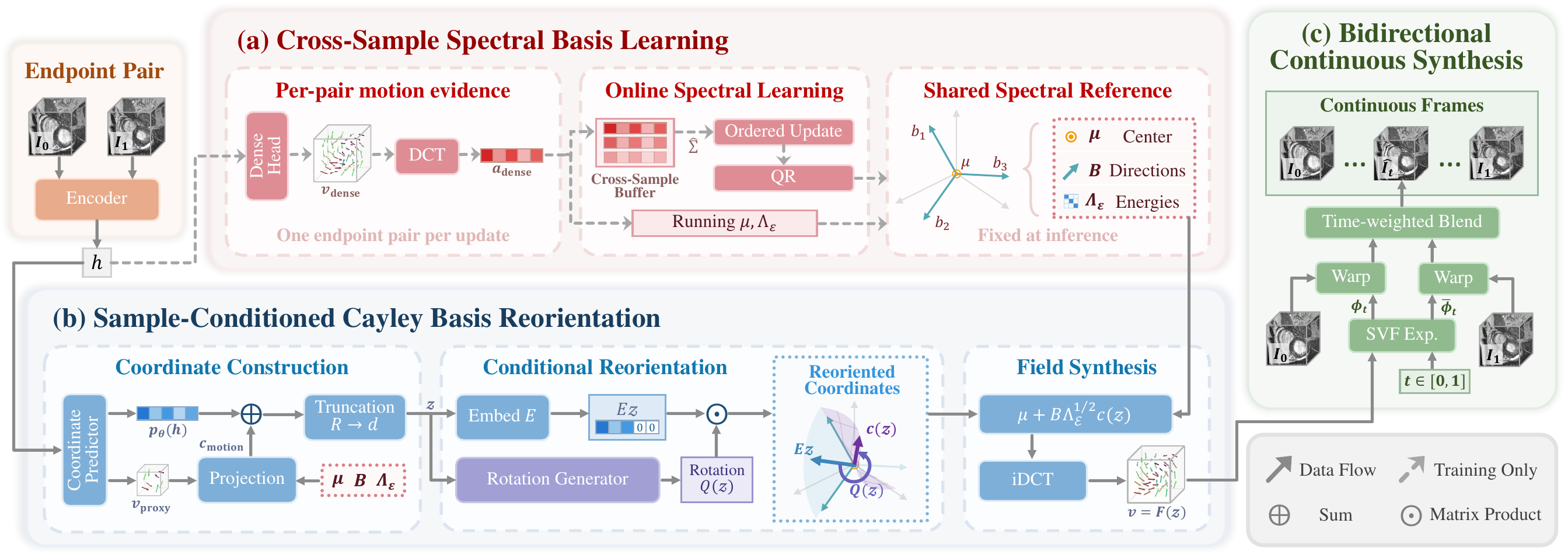}
\caption{Overview of the proposed framework for unsupervised 4D medical image interpolation. (a) Endpoint-derived dense estimates are aggregated in the DCT domain to learn a shared spectral reference $(\mu,B,\Lambda_\epsilon)$. (b) Predicted motion coordinates condition the reorientation of shared directions and their combination into a complete velocity field. (c) Velocity integration, endpoint warping, and time-weighted blending generate intermediate volumes. The shared reference is fixed at inference; dashed arrows indicate training-only paths.}
\label{fig:overview}
\end{figure*}

\section{Related Work}

\subsection{4D Medical Image Interpolation}

Research on 4D medical image interpolation has developed motion representations and synthesis models for recovering unobserved anatomical changes~\cite{guo2020spatiotemporal,kim2024data}. More broadly, joint motion estimation and image reconstruction exploit the dependence of recovered detail on correspondence quality~\cite{zhi2023coarse,pan2024motion}. Among image interpolation methods, motion-guided approaches such as SVIN and MPVF use multi-scale representations to accommodate different motion magnitudes~\cite{guo2020spatiotemporal,wei2023mpvf}. AMT strengthens matching through all-pairs correspondence and multiple motion fields~\cite{li2023amt}, while VFIFME reduces reconstruction's dependence on imperfect motion estimates~\cite{yu2023range}. Generative modeling supplies temporal priors through deformation-aware diffusion in DDM~\cite{kim2022diffusion}, latent dynamics in LDDM~\cite{chen2024ultrasound}, and Fourier guidance in FB-Diff~\cite{you2025fb}. These priors complement correspondence-based synthesis by drawing on recurring patterns of anatomical appearance and change.

The supervision used to learn motion varies across approaches. TMSDF transfers intermediate motion cues into time-modulated deformation fusion~\cite{zhang2025temporal}. For unsupervised interpolation without intermediate training targets, UVI-Net uses cyclic endpoint reconstruction~\cite{kim2024data}, while TTT4MII uses instance-specific self-supervision~\cite{zhang2025test}. Temporal parameterization and spatial consistency also influence the recovered anatomical sequence~\cite{li2026phase,zhou2025diffusion}. These designs strengthen reconstruction and temporal modeling, although endpoint agreement alone does not determine the intervening anatomical motion.

\subsection{Motion Ambiguity in Deformation Modeling}

Motion ambiguity arises when different motion fields explain the same observed anatomy~\cite{reithmeir2026regularization}. Spatial regularization in VM and TM~\cite{balakrishnan2019voxelmorph,chen2022transmorph} and velocity integration~\cite{beg2005computing,vercauteren2009diffeomorphic} guide deformation through local coherence and transformation structure. Shape-aware tracking uses anatomical boundaries to constrain correspondence~\cite{meng2022mulvimotion}; physical formulations incorporate tissue mechanics~\cite{arriatalopez2023warppinn} or signal consistency~\cite{hanania2025mbss} to supplement image agreement. Fourier-Net+~\cite{jia2026fourier} and LRVF~\cite{li2026low} restrict motion variation through band-limited and low-rank representations, with coefficient-based respiratory models providing related constraints~\cite{huttinga2021real,huang2025digs}. Spatial penalties couple neighboring displacements, while compact coefficients link distant regions through shared directions. These restrictions narrow the search but do not specify every unobserved anatomical state. Recent work examines non-unique motion~\cite{zhou2023exploring}, including velocity ambiguity in VD~\cite{zhong2026velocity}. Probabilistic motion formulations model deformation distributions under partial observations~\cite{thibeault2024conditional,huttinga2023gaussian}.

A learned motion manifold restricts candidate fields through coordinates describing their coordinated variation~\cite{meila2024manifold}. Statistical respiratory models~\cite{ehrhardt2011statistical} and learned deformation representations~\cite{krebs2019learning,mezheritsky2022population} extract recurring structure; manifold regularization guides estimated fields toward a learned family~\cite{bhalodia2019cooperative}. Related cardiac representations encode shape and temporal variation~\cite{rohe2018lowdimensional,dou2025cardiosynth,qiao2025mesh4d}, with conditional neural fields relating geometry to individual characteristics~\cite{sorensen2024spatiotemporal}. Case-specific neural fields~\cite{sun2024medical,shen2024continuous,gebauer2026nonlinear} and anatomy-conditioned motion modes~\cite{zhu2026learning} make the representation responsive to individual observations. The shared family provides a reusable constraint across cases; conditioning or case-wise fitting determines which combinations of this structure are available for each observed anatomy. For unsupervised interpolation, the challenge is to learn a shared motion manifold from endpoint-derived evidence while retaining the motion needed by each pair.

\section{Methodology}

\subsection{Problem Formulation and Overview}
\label{sec:problem_overview}

Let $I_0,I_1:\Omega\rightarrow\mathbb R$ denote two observed volumes on a spatial domain $\Omega\subset\mathbb R^3$, with their acquisition times normalized to $0$ and $1$. Unsupervised interpolation estimates $\widehat I_t$ for any $t\in(0,1)$ without intermediate-volume supervision. We represent motion by a stationary velocity field (SVF) $v$, whose integrated deformations reconstruct the endpoints and synthesize intermediate volumes through backward warping, $\mathcal W(I,\phi)=I\circ\phi$. Endpoint reconstruction supplies the training signal, but distinct velocity fields can explain the same observations and produce different intermediate anatomy.

Our framework learns a motion manifold that restricts the candidate velocity fields:
\begin{equation}
\mathcal M=\{F(z):z\in\mathcal Z\subset\mathbb R^d\},
\label{eq:motion_family}
\end{equation}
where $z$ contains the intrinsic motion coordinates and $F$ maps them to a complete 3D velocity field. As illustrated in Fig.~\ref{fig:overview}, Cross-Sample Spectral Basis Learning extracts shared motion structure from endpoint-derived estimates, and Sample-Conditioned Cayley Basis Reorientation determines how this structure is expressed for the current endpoint pair. The resulting family limits independent motion variation while accommodating sample-specific directions.

\subsection{Cross-Sample Spectral Basis Learning}
\label{sec:spectral_basis_learning}

Shared motion structure must be learned from endpoint observations, since intermediate motion is unavailable. A shared 3D encoder and an auxiliary prediction head estimate a dense SVF $v_{\mathrm{dense}}$ from each training endpoint pair. Bidirectional endpoint reconstruction and spatial regularization train this head~\cite{balakrishnan2019voxelmorph}, providing provisional motion estimates across different anatomical states. Across training pairs, their recurring, spatially coordinated variation supplies the evidence for learning the shared basis.

We apply an orthonormal discrete cosine transform (DCT)~\cite{wang2023dct} to each velocity component and retain a low-frequency block, giving $a_{\mathrm{dense}}=\mathcal C(v_{\mathrm{dense}})\in\mathbb R^M$, where $M=3K_DK_HK_W$ and $(K_D,K_H,K_W)$ are the retained block dimensions. The transpose operator $\mathcal C^{\top}$ performs the inverse DCT (iDCT) after zero-padding the omitted coefficients. Within this spectral representation, an ordered basis $B=[b_1,\ldots,b_R]\in\mathbb R^{M\times R}$, with $B^{\top}B=I_R$ and $R<M$, is learned online from centered coefficients. For their buffered covariance $\widehat\Sigma$, we use the update direction:
\begin{equation}
\Delta B=\widehat\Sigma B-B\operatorname{triu}(B^{\top}\widehat\Sigma B),
\label{eq:shared_basis_update}
\end{equation}
where $\operatorname{triu}(\cdot)$ retains the upper-triangular entries. The step along $\Delta B$ is normalized by the centered buffered coefficients' mean squared norm and followed by QR orthonormalization. This ordered covariance update~\cite{sanger1989optimal} accumulates dominant variation across endpoint pairs while the dense estimates evolve during training.

Let $\mu$ denote the running coefficient mean, and let the positive diagonal matrix $\Lambda_{\epsilon}$ contain the running mean squared centered projections along the basis directions, floored at $\epsilon$ for numerical normalization. Using the shared spectral reference $(\mu,B,\Lambda_\epsilon)$, we project the dense estimate into motion coordinates and reconstruct its velocity field as follows:
\begin{equation}
\begin{aligned}
c_{\mathrm{dense}}&=\Lambda_{\epsilon}^{-1/2}B^{\top}(a_{\mathrm{dense}}-\mu),\\
v_{\mathrm{proj}}&=\mathcal C^{\top}(\mu+B\Lambda_{\epsilon}^{1/2}c_{\mathrm{dense}}).
\end{aligned}
\label{eq:shared_motion_projection}
\end{equation}

The motion coordinate $c_{\mathrm{dense}}\in\mathbb R^R$ and its reconstructed field $v_{\mathrm{proj}}$ provide coordinate- and velocity-level targets for the sample-specific representation. Each learned direction can combine retained frequencies and velocity components, so a change in one motion coordinate describes coordinated variation in a complete field. These targets express the endpoint-derived motion evidence through the shared basis, linking the motion needed by an endpoint pair to patterns supported across training cases.

\subsection{Sample-Conditioned Cayley Basis Reorientation}
\label{sec:cayley_basis_reorientation}

The shared reference constrains the available motion patterns, while each endpoint pair determines their sample-specific expression. We construct $d$ intrinsic motion coordinates from the shared encoder features $h$:
\begin{equation}
z=[p_{\theta}(h)+c_{\mathrm{motion}}]_{1:d}\in\mathbb R^d,
\label{eq:intrinsic_motion_prediction}
\end{equation}
where $p_{\theta}(h)\in\mathbb R^R$ is a learned coordinate prediction, $[\cdot]_{1:d}$ truncates the summed coordinates to the leading $d<R$ components, and $c_{\mathrm{motion}}\in\mathbb R^R$ supplements the prediction with a motion-derived cue. This cue applies the same centered, energy-normalized spectral projection to an upsampled coarse velocity proxy $v_{\mathrm{proxy}}$ produced within the coordinate predictor.

To let the same number of coordinates express different motion directions across samples, we embed $z$ into $\mathbb R^R$ using a fixed matrix $E\in\mathbb R^{R\times d}$ that appends $R-d$ zeros. An MLP $G_{\psi}$ takes $z$ as input and predicts two factors $U(z),V(z)\in\mathbb R^{R\times s}$, which define the skew-symmetric generator:
\begin{equation}
K(z)=\frac{U(z)V(z)^{\top}-V(z)U(z)^{\top}}{\sqrt{sR}}.
\label{eq:skew_generator}
\end{equation}

\begin{figure}[!t]
\centering
\includegraphics[width=\columnwidth]{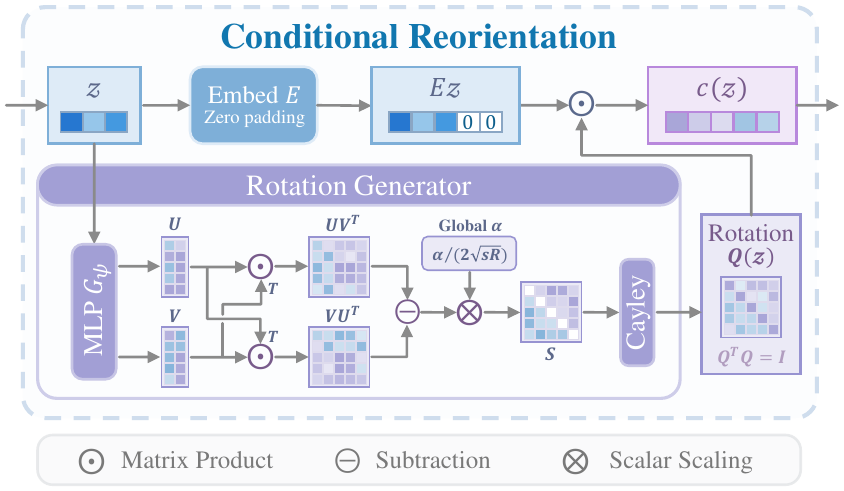}
\caption{Conditional reorientation of motion coordinates. The motion coordinates $z$ generate a Cayley rotation $Q(z)$ that acts on their embedding $Ez$. This changes the effective motion directions for each pair while preserving coordinate length and the number of independent coordinates.}
\label{fig:cayley_detail}
\end{figure}

A bounded scalar gate $\alpha=g_{\max}\tanh(\rho)$ scales the reorientation generator. With $S(z)=\alpha K(z)/2$, the Cayley map in Fig.~\ref{fig:cayley_detail} gives:
\begin{equation}
Q(z)=(I_R-S(z))^{-1}(I_R+S(z)).
\label{eq:cayley_reorientation}
\end{equation}

Skew-symmetry makes $Q(z)$ orthogonal, preserving the length of the normalized motion coordinates~\cite{helfrich2018orthogonal}. The factorization restricts its departure from the identity to rank at most $2s$. Initializing the learnable scalar $\rho$ at zero starts the model with the unrotated embedding, from which sample-conditioned reorientation is learned.

The reoriented motion coordinate is:
\begin{equation}
c(z)=Q(z)Ez.
\label{eq:reoriented_motion_coordinate}
\end{equation}

Although $c(z)$ can use all $R$ components, each remains determined by the same $d$ intrinsic coordinates. Its combination with the shared basis yields the motion map:
\begin{equation}
F(z)=\mathcal C^{\top}\!\left(\mu+B\Lambda_{\epsilon}^{1/2}c(z)\right).
\label{eq:motion_synthesis}
\end{equation}

The columns of $B\Lambda_{\epsilon}^{1/2}Q(z)E$ specify the effective spectral directions for velocity synthesis. A fixed embedding would use the same $d$-dimensional span for every pair. Reorientation allows different pairs to draw on different combinations of the $R$ shared directions. The same $d$ coordinates determine both their combination and orientation, allowing the motion family to accommodate individual anatomy and motion state without additional independently adjustable per-pair variables.

\subsection{Theoretical Analysis of Motion Ambiguity}
\label{sec:theoretical_analysis}

Locally, the benefit of restricting motion to the learned manifold depends on how its coordinates affect endpoint reconstruction. For scalar volumes with $N$ voxels, let $g(z)\in\mathbb R^{2N}$ concatenate the residuals $\mathcal W(I_0,\exp(F(z)))-I_1$ and $\mathcal W(I_1,\exp(-F(z)))-I_0$, where $\mathcal W$ denotes image warping and $\exp$ integrates an SVF. Its coordinate sensitivity at $z_{\star}$ is $H_{\star}=Dg(z_{\star})$.

\noindent\textbf{Proposition 1 (Local identifiability of motion coordinates).} Assume that $F$ and the endpoint-warping residual are continuously differentiable locally. The paired residual of an unrestricted $v\in\mathbb R^{3N}$ has at least $N$ first-order null directions. If $H_{\star}$ has full column rank $d$, then $z_{\star}$ is locally identifiable from the paired residual and $F$ is locally an immersion. If $Dg$ is also $L$-Lipschitz, with $L>0$, and $\kappa_{\star}=\sigma_{\min}(H_{\star})>0$, the following bound holds:
\begin{equation}
\lVert g(z_{\star}+\delta z)-g(z_{\star})\rVert_2\geq\frac{\kappa_{\star}}{2}\lVert\delta z\rVert_2,
\label{eq:local_identifiability_bound}
\end{equation}
for $\lVert\delta z\rVert_2\leq\kappa_{\star}/L$ whose connecting segment stays within the neighborhood.

\noindent\textbf{Proof.} The dense residual Jacobian has $2N$ rows and $3N$ columns, so rank--nullity gives a nullity of at least $N$. Full column rank of $H_{\star}$ provides a nonsingular $d\times d$ row minor; the inverse-function theorem applied to those residual components gives local injectivity. The chain rule also implies that $DF(z_{\star})$ has rank $d$. Finally, the first-order remainder is bounded by $L\lVert\delta z\rVert_2^2/2$. Subtracting it from $\lVert H_{\star}\delta z\rVert_2\geq\kappa_{\star}\lVert\delta z\rVert_2$ yields the bound.

The result has two implications. 1) Freely varying dense fields contain motion changes to which endpoint reconstruction is insensitive at first order. 2) Within the learned family, full column rank means that every local coordinate direction affects the endpoint residual. The bound quantifies this sensitivity: sufficiently small coordinate changes produce a proportionate residual change. Restricting the motion family can therefore make nearby motions distinguishable when its retained coordinates are observable from the endpoints.

\begin{table*}[t]
\centering
\caption{Quantitative comparison on ACDC and 4D-Lung. Results are mean $\pm$ standard deviation over five training runs. Sup. indicates intermediate-volume supervision. Model size (\#P) is also reported. Best reconstruction scores are in bold.}
\label{tab:comparison}
\footnotesize
\setlength{\tabcolsep}{2pt}
\renewcommand{\arraystretch}{1.1}

\begin{tabular*}{\textwidth}{@{\extracolsep{\fill}}lcc*{6}{c}r@{}}
\toprule
\tworowcell{Method} & \tworowcell{Venue} & \tworowcell{Sup.} & \multicolumn{3}{c}{ACDC}
& \multicolumn{3}{c}{4D-Lung}
& \tworowcell{\#P (M)} \\
\cmidrule(lr){4-6} \cmidrule(lr){7-9}
 & & & PSNR $_{\text{dB}}^{\uparrow}$ & NMI $_{\times 10^{-2}}^{\uparrow}$ & SSIM $_{\times 10^{-2}}^{\uparrow}$
& PSNR $_{\text{dB}}^{\uparrow}$ & NMI $_{\times 10^{-2}}^{\uparrow}$ & SSIM $_{\times 10^{-2}}^{\uparrow}$ & \\
\midrule
VM~\cite{balakrishnan2019voxelmorph} & TMI'19 & \ding{55}
& 25.694 $_{\pm 0.074}$ & 53.543 $_{\pm 0.367}$ & 94.678 $_{\pm 0.104}$
& 26.077 $_{\pm 0.066}$ & 43.982 $_{\pm 0.355}$ & 86.055 $_{\pm 0.292}$ & 10.044 \\
TM~\cite{chen2022transmorph} & MedIA'22 & \ding{55}
& 27.125 $_{\pm 0.200}$ & 64.628 $_{\pm 0.671}$ & 96.644 $_{\pm 0.123}$
& 26.683 $_{\pm 0.051}$ & 48.945 $_{\pm 0.104}$ & 84.779 $_{\pm 0.125}$ & 10.804 \\
SVIN~\cite{guo2020spatiotemporal} & CVPR'20 & \ding{55}
& 28.434 $_{\pm 0.443}$ & 60.178 $_{\pm 0.335}$ & 97.352 $_{\pm 0.152}$
& 27.312 $_{\pm 0.280}$ & 43.171 $_{\pm 0.392}$ & 85.442 $_{\pm 0.286}$ & 12.393 \\
UVI-Net~\cite{kim2024data} & CVPR'24 & \ding{55}
& 31.903 $_{\pm 0.118}$ & 72.252 $_{\pm 0.166}$ & 98.618 $_{\pm 0.027}$
& 28.948 $_{\pm 0.277}$ & 55.157 $_{\pm 0.841}$ & 88.750 $_{\pm 0.668}$ & 14.273 \\
TTT4MII~\cite{zhang2025test} & IJCNN'25 & \ding{55}
& 32.136 $_{\pm 0.140}$ & 72.514 $_{\pm 0.174}$ & 98.671 $_{\pm 0.031}$
& 28.322 $_{\pm 0.529}$ & 50.590 $_{\pm 1.563}$ & 83.495 $_{\pm 1.280}$ & 12.006 \\
Fourier-Net+~\cite{jia2026fourier} & TNNLS'26 & \ding{55}
& 28.815 $_{\pm 0.072}$ & 65.098 $_{\pm 0.039}$ & 97.290 $_{\pm 0.024}$
& 26.782 $_{\pm 0.191}$ & 49.032 $_{\pm 0.194}$ & 84.814 $_{\pm 0.424}$ & 12.414 \\
LRVF~\cite{li2026low} & MICCAI'26 & \ding{55} & 31.961 $_{\pm 0.168}$ & 72.610 $_{\pm 0.138}$ & 98.623 $_{\pm 0.028}$ & 29.823 $_{\pm 0.106}$ & 62.337 $_{\pm 0.205}$ & 91.286 $_{\pm 0.126}$ & 33.917 \\
\midrule
MPVF~\cite{wei2023mpvf} & JBHI'23 & \ding{51}
& 32.144 $_{\pm 0.011}$ & 70.658 $_{\pm 0.038}$ & 98.669 $_{\pm 0.003}$
& \textbf{29.909} $_{\pm 0.035}$ & 53.074 $_{\pm 0.389}$ & 90.180 $_{\pm 0.053}$ & 14.533 \\
LDDM~\cite{chen2024ultrasound} & MICCAI'24 & \ding{51}
& 31.839 $_{\pm 0.033}$ & 72.333 $_{\pm 0.302}$ & 98.551 $_{\pm 0.013}$
& 27.563 $_{\pm 0.111}$ & 56.577 $_{\pm 8.077}$ & 83.499 $_{\pm 12.092}$ & 15.545 \\

AMT~\cite{li2023amt} & CVPR'23 & \ding{51}
& 31.009 $_{\pm 0.267}$ & 71.461 $_{\pm 0.438}$ & 98.389 $_{\pm 0.077}$
& 28.881 $_{\pm 0.773}$ & 62.101 $_{\pm 1.080}$ & 87.954 $_{\pm 1.723}$ & 23.336\\
VFIFME~\cite{yu2023range} & CVPR'23 & \ding{51}
& 29.627 $_{\pm 3.297}$ & 66.215 $_{\pm 8.075}$ & 97.250 $_{\pm 4.624}$
& 29.711 $_{\pm 1.183}$ & 50.843 $_{\pm 4.208}$ & 89.677 $_{\pm 1.887}$ & 24.089 \\
FB-Diff~\cite{you2025fb} & ICCV'25 & \ding{51}
& 29.387 $_{\pm 0.103}$ & 62.135 $_{\pm 0.274}$ & 97.483 $_{\pm 0.163}$
& 27.617 $_{\pm 0.867}$ & 49.275 $_{\pm 1.345}$ & 80.913 $_{\pm 3.712}$ & 33.527 \\
DDM~\cite{kim2022diffusion} & MICCAI'22 & \ding{51}
& 28.558 $_{\pm 1.679}$ & 64.142 $_{\pm 1.349}$ & 97.237 $_{\pm 0.872}$
& 26.896 $_{\pm 0.101}$ & 42.544 $_{\pm 0.894}$ & 85.229 $_{\pm 0.270}$ & 23.178 \\
TMSDF~\cite{zhang2025temporal} & MICCAI'25 & \ding{51}
& 31.974 $_{\pm 0.225}$ & 66.401 $_{\pm 0.833}$ & 98.558 $_{\pm 0.072}$
& 29.848 $_{\pm 0.023}$ & 48.360 $_{\pm 0.113}$ & 90.289 $_{\pm 1.086}$ & 42.283 \\
VD~\cite{zhong2026velocity} & TPAMI'26 & \ding{51}
& 32.107 $_{\pm 0.962}$ & 72.136 $_{\pm 2.612}$ & 98.680 $_{\pm 0.368}$
& 28.982 $_{\pm 0.877}$ & 60.657 $_{\pm 1.443}$ & 90.150 $_{\pm 1.135}$ & 23.336 \\
\midrule
\textbf{Ours} & - & \ding{55}
& \textbf{32.205} $_{\pm 0.004}$ & \textbf{72.734} $_{\pm 0.001}$ & \textbf{98.685} $_{\pm 0.001}$
& 29.874 $_{\pm 0.017}$ & \textbf{62.442} $_{\pm 0.068}$ & \textbf{91.302} $_{\pm 0.028}$ & 15.100 \\
\bottomrule
\end{tabular*}
\end{table*}

\subsection{Motion Reconstruction and Learning}
\label{sec:motion_reconstruction_learning}

For a velocity field $u$, $\mathcal L_{\mathrm{rec}}(u)$ sums bidirectional endpoint reconstruction penalties combining Charbonnier intensity discrepancy and local normalized cross-correlation~\cite{kim2024data}. The dense objective $\mathcal L_{\mathrm{dense}}$ adds mean squared spatial velocity gradients with weight $\lambda_r$ to $\mathcal L_{\mathrm{rec}}(v_{\mathrm{dense}})$, while $\mathcal L_{\mathrm{man}}=\mathcal L_{\mathrm{rec}}(v)$ trains the manifold prediction $v=F(z)$. Projection, velocity, and coordinate consistency couple these predictions to the shared reference:
\begin{equation}
\begin{gathered}
\mathcal L_{\mathrm{proj}}=\lVert v_{\mathrm{dense}}-v_{\mathrm{proj}}\rVert_1,\\
\mathcal L_{\mathrm{vel}}=\lVert v-v_{\mathrm{proj}}\rVert_1,\quad
\mathcal L_{\mathrm{coord}}=\lVert c(z)-c_{\mathrm{dense}}\rVert_1,
\end{gathered}
\label{eq:motion_consistency_losses}
\end{equation}
where $\lVert\cdot\rVert_1$ denotes mean absolute discrepancy. The full objective is:
\begin{equation}
\mathcal L=\mathcal L_{\mathrm{dense}}+\mathcal L_{\mathrm{man}}+\lambda_p\mathcal L_{\mathrm{proj}}+\lambda_v\mathcal L_{\mathrm{vel}}+\lambda_c\mathcal L_{\mathrm{coord}}.
\label{eq:total_objective}
\end{equation}

The projection term keeps the dense estimates compatible with the shared basis. Velocity and coordinate consistency use the projected field and coordinates as fixed targets to guide sample-specific prediction at field and coordinate levels, respectively.

\noindent\textbf{Inference.} With the basis and running statistics fixed, the coordinate predictor estimates one SVF $v=F(z)$ for the endpoint pair from shared encoder features, without the auxiliary dense head. For query time $t$, scaling-and-squaring~\cite{vercauteren2009diffeomorphic} gives $\phi_t=\exp(tv)$ and $\bar\phi_t=\exp(-(1-t)v)$, and bidirectional fusion yields:
\begin{equation}
\widehat I_t=(1-t)\mathcal W(I_0,\phi_t)+t\mathcal W(I_1,\bar\phi_t).
\label{eq:bidirectional_reconstruction}
\end{equation}

\section{Experiments}

\subsection{Experimental Settings}

\noindent\textbf{Datasets.} We evaluate on ACDC~\cite{bernard2018deep} and 4D-Lung~\cite{hugo2016data}. ACDC contains 150 cardiac MRI cases, divided into 80/20/50 cases for training/validation/testing. End-diastolic and end-systolic volumes serve as endpoints, and volumes are resized to $160\times160\times16$. Our processed 4D-Lung subset contains 500 respiratory samples, split at the patient level into 306/84/110 samples for training/validation/testing. The $0\%$ and $50\%$ respiratory phases serve as endpoints, with volumes resized to $128\times128\times32$. All volumes are histogram-equalized and linearly normalized to $[0,1]$.

\noindent\textbf{Evaluation Protocol.} Fifteen baselines cover deformation-based modeling, motion-guided interpolation, and generative synthesis under common data partitions, preprocessing, and test volumes. Table~\ref{tab:comparison} identifies intermediate-volume supervision. Our training losses use only endpoints; intervening volumes are reserved for validation-based model selection and evaluation. Reconstruction is evaluated at recorded acquisition times using PSNR, NMI, and SSIM~\cite{wang2004image}, with scores aggregated over test volumes and standard deviations based on five training runs.

\noindent\textbf{Implementation Details.} The PyTorch model uses a shared four-stage 3D convolutional encoder with channel widths $40/80/160/320$, an auxiliary dense head, and a motion-coordinate predictor, and is trained on NVIDIA RTX A6000 GPUs. The retained DCT block is $8\times32\times32$, with $R=80$, $d=40$, Cayley factor width $s=4$, and gate bound $g_{\max}=0.25$. We set $\lambda_r=0.01$, $\lambda_p=10$, $\lambda_v=1$, and $\lambda_c=5$, with relative weights $10:1$ for the Charbonnier and correlation penalties. AdamW uses learning rates of $10^{-3}$ for the dense branch and $3\times10^{-4}$ for the remaining components, with cosine annealing. Training runs for 500 epochs on ACDC and 200 on 4D-Lung, using batch size 1 and gradient accumulation over eight iterations. Velocity integration uses seven scaling-and-squaring steps.

\begin{figure*}[t]
\centering
\includegraphics[width=\textwidth]{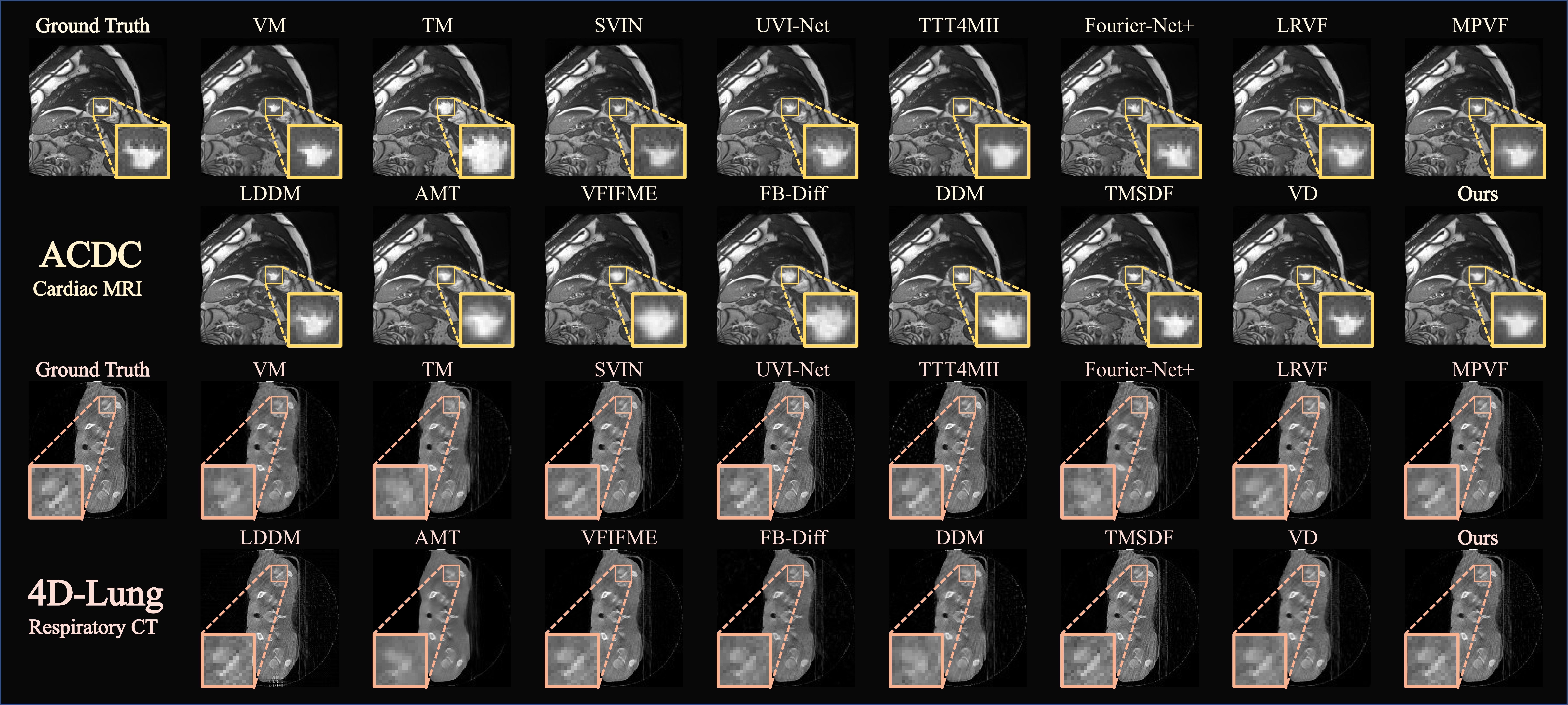}
\caption{Qualitative comparison on ACDC and 4D-Lung. Enlarged regions highlight the ventricular blood-pool boundary and internal low-signal structures in ACDC, and fine elongated intrapulmonary structures in 4D-Lung.}
\label{fig:qualitative_comparison}
\end{figure*}

\begin{figure*}[!t]
\centering
\includegraphics[width=\textwidth]{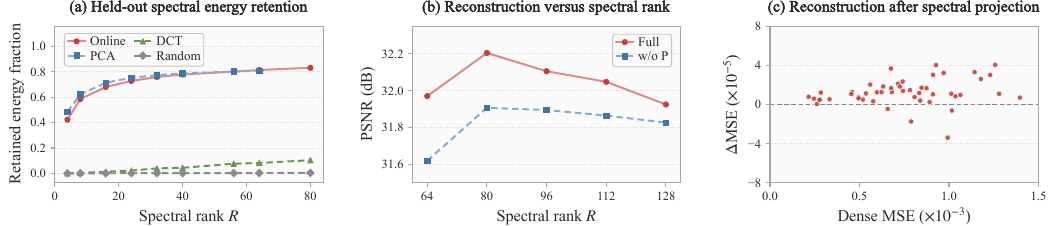}
\caption{Shared-basis analyses on ACDC. (a) Retained energy fraction for learned, PCA, frequency-ordered DCT, and random directions. (b) Reconstruction versus $R$ at $d=40$, with and without projection consistency $P$. (c) Change in intermediate-volume MSE after projecting the dense field, plotted against its original MSE; positive $\Delta\mathrm{MSE}$ favors projection.}
\label{fig:spectral_basis_analysis}
\end{figure*}

\begin{figure*}[t]
\centering
\includegraphics[width=\textwidth]{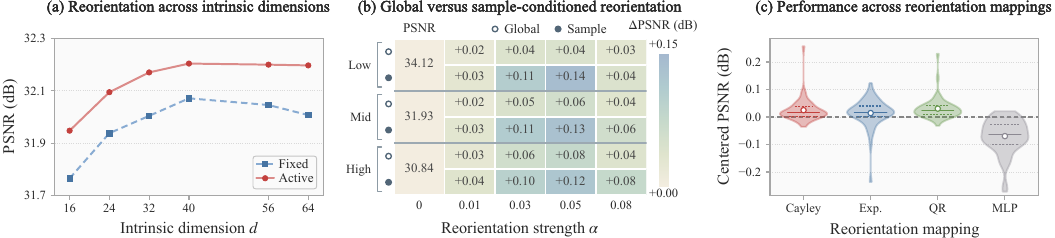}
\caption{Reorientation analyses on ACDC. (a) Fixed embedding versus sample-conditioned reorientation at $R=80$. (b) Global versus sample-conditioned reorientation across low, medium, and high image-change groups at $d=40$. The $\alpha=0$ column gives absolute PSNR; remaining columns show gains over this reference. (c) Cayley, matrix-exponential, QR-based, and normalized-MLP mappings, with PSNR expressed as deviations from each case's mean across the four mappings.}
\label{fig:reorientation_analysis}
\end{figure*}

\subsection{Quantitative Comparison}
\label{sec:comparison}

Table~\ref{tab:comparison} contrasts methods with different sources of motion guidance. Cyclic reconstruction in UVI-Net~\cite{kim2024data} and instance-specific refinement in TTT4MII~\cite{zhang2025test} substantially improve ACDC reconstruction over VM, TM, and Fourier-Net+. Their less uniform gains on 4D-Lung, particularly across NMI and SSIM, underline the difficulty of inferring intervening anatomy from endpoint supervision alone. Our method maintains strong performance across all three criteria on both datasets. Cardiac contraction and respiratory deformation produce different patterns of image change; consistent results in both settings support a data-derived motion constraint across these distinct anatomical dynamics. Agreement across intensity, statistical dependence, and structural similarity measures also provides complementary evidence of reconstruction fidelity.

Intermediate-supervised methods provide a complementary comparison. Multi-scale deformation fusion in MPVF and TMSDF~\cite{wei2023mpvf,zhang2025temporal}, multiple motion fields in AMT~\cite{li2023amt}, and distance-indexed guidance in VD~\cite{zhong2026velocity} enrich the motion information used for synthesis. Their strengths differ across criteria: on 4D-Lung, MPVF achieves high PSNR and AMT high NMI, while their SSIM is less competitive. Intermediate supervision enriches the training cues, but reconstruction still depends on how these cues organize the estimated motion. Our method exceeds most of these approaches and remains close to the strongest results. Its near-leading respiratory PSNR is accompanied by strong NMI and SSIM, showing that similar overall intensity error can coexist with different levels of structural agreement. Achieving this balance without intermediate training targets supports the shared motion constraint as useful interpolation guidance.

\subsection{Qualitative Comparison}

As shown in Fig.~\ref{fig:qualitative_comparison}, the methods recover broadly similar anatomy, but differ in how local boundaries and fine structures are reconstructed. In ACDC, our result retains the indentations along the ventricular blood-pool boundary and the separation of internal low-signal structures, which become smoother or less distinct in several predictions. On 4D-Lung, our reconstruction preserves narrow intrapulmonary profiles close to the reference. These differences reflect the spatial arrangement of local structures as well as their sharpness. A plausible-looking boundary may still have the wrong extent, while averaging misaligned structures can broaden or merge a narrow profile. The enlarged regions show that the learned motion family preserves this localized variation alongside the overall anatomical change.

\subsection{Analysis of Spectral Basis Learning}
\label{sec:spectral_basis_analysis}

\noindent\textbf{Spectral energy retention.} Fig.~\ref{fig:spectral_basis_analysis}(a) compares retained energy fractions, $\lVert B^{\top}(a_{\mathrm{dense}}-\mu)\rVert_2^2/\lVert a_{\mathrm{dense}}-\mu\rVert_2^2$, averaged over held-out cases~\cite{halko2011finding}. PCA is fitted to training dense-motion coefficients; frequency-ordered DCT and random orthonormal directions provide fixed alternatives in the same DCT block. The learned directions approach PCA and substantially exceed these alternatives at matched $R$. Their alignment with recurring motion variation uses the shared direction budget more effectively than frequency order alone.

\noindent\textbf{Reconstruction versus spectral rank.} Increasing $R$ at fixed $d=40$ first improves reconstruction and then reduces it, with a peak near $R=80$ in Fig.~\ref{fig:spectral_basis_analysis}(b). Here $P$ denotes projection consistency $\mathcal L_{\mathrm{proj}}$. Since $d$ is unchanged, this comparison varies the breadth of the shared reference while preserving the number of coordinates assigned to each pair. The trend distinguishes coverage of dense variation from useful interpolation capacity: a wider set of available directions does not necessarily improve the intermediate anatomy. The consistently higher scores with $P$ further indicate the value of keeping the evolving dense evidence aligned with the shared representation.

\noindent\textbf{Reconstruction after spectral projection.} For each endpoint pair, we synthesize intermediate volumes using the dense field and its projection through the same integration and blending procedure. Each point in Fig.~\ref{fig:spectral_basis_analysis}(c) reports $\Delta\mathrm{MSE}=\mathrm{MSE}_{\mathrm{dense}}-\mathrm{MSE}_{\mathrm{proj}}$, averaged over that case's intermediate volumes. The small differences and predominance of positive values show that projection largely preserves reconstruction-relevant information through velocity integration and image synthesis. Energy retention measures coverage of dense variation, while this reconstruction-level check evaluates restriction after integration and warping. Positive differences also occur for cases with larger dense reconstruction errors. The selected directions retain variation that remains useful through the nonlinear synthesis process, helping explain why projected fields preserve or improve intermediate reconstruction.

\subsection{Analysis of Sample-Conditioned Reorientation}
\label{sec:reorientation_analysis}

\noindent\textbf{Reorientation across intrinsic dimensions.} Fixed uses the unrotated embedding $c=Ez$, while Active uses $c=Q(z)Ez$, with other settings held fixed. Active improves reconstruction throughout the tested range of $d$ in Fig.~\ref{fig:reorientation_analysis}(a), and its performance levels off near $d=40$. A smaller Active coordinate set can also match or exceed larger Fixed configurations. The advantage reflects how effectively the shared directions are used as well as the number of independent motion variables. The fixed embedding gives every pair the same motion directions, even though the anatomy and motion state differ. Sample-conditioned reorientation preserves useful individual variation within a restricted coordinate budget, reducing the need to recover expression solely by enlarging that budget.

\noindent\textbf{Global versus sample-conditioned reorientation.} Global learns one orientation shared by all cases, whereas Sample predicts the orientation from each pair's motion coordinates; other settings are held fixed. Cases are ranked by mean absolute foreground intensity change between sampled reference volumes and divided into low, medium, and high thirds. In Fig.~\ref{fig:reorientation_analysis}(b), Sample yields larger gains at matched generator strengths $\alpha$, with both variants returning to the unrotated embedding at $\alpha=0$. The consistent advantage favors adapting the shared directions to each endpoint pair. Even the low-change group benefits, indicating that sample dependence is useful beyond cases with large image changes. Moderate generator strengths give the largest gains, balancing directional flexibility with controlled departures from the unrotated shared reference.

\noindent\textbf{Alternative reorientation mappings.} We compare Cayley with matrix-exponential and QR-based orthogonal mappings, and with an MLP that directly predicts length-normalized coordinates. To remove between-case difficulty offsets, each PSNR is centered by subtracting that case's mean across the four mappings. The distributions in Fig.~\ref{fig:reorientation_analysis}(c) place Cayley close to the other orthogonal constructions and above the normalized MLP. The MLP's lower scores favor structured directional mappings over directly predicting a length-matched coordinate vector. The comparable performance of the orthogonal variants supports structured directional adaptation as the central benefit of reorientation. Cayley provides a competitive realization of this flexibility with norm preservation and a low-rank-controlled departure from the shared directions.

\subsection{Additional Experiments}

\noindent\textbf{Loss configuration.} Table~\ref{tab:loss_ablation} compares combinations of $\mathcal L_{\mathrm{proj}}$, $\mathcal L_{\mathrm{vel}}$, and $\mathcal L_{\mathrm{coord}}$. Projection consistency improves the base configuration, while coupling it with the sample-specific consistency terms yields further gains. Combining these terms brings the shared structure into both dense motion estimation and sample-specific prediction. Velocity consistency ties the guidance to the field used for image synthesis, while coordinate consistency aligns the compact representation. Their joint advantage suggests that constraining the final field and organizing its predicted coordinates play complementary roles in interpolation.

\begin{table}[t]
\centering
\caption{Loss ablation on ACDC.}
\label{tab:loss_ablation}
\footnotesize
\setlength{\tabcolsep}{3pt}
\renewcommand{\arraystretch}{1.1}
\begin{tabular}{@{}ccc*{3}{c}@{}}
\toprule
$\mathcal L_{\mathrm{proj}}$ & $\mathcal L_{\mathrm{vel}}$ & $\mathcal L_{\mathrm{coord}}$ & PSNR $_{\text{dB}}^{\uparrow}$ & NMI $_{\times 10^{-2}}^{\uparrow}$ & SSIM $_{\times 10^{-2}}^{\uparrow}$ \\
\midrule
 & & & 30.855 & 66.625 & 98.295 \\
\ding{51} & & & 31.520 & 68.754 & 98.507 \\
 & \ding{51} & \ding{51} & 31.907 & 72.416 & 98.614 \\
\ding{51} & & \ding{51} & 32.030 & 72.672 & 98.645 \\
\ding{51} & \ding{51} & & 32.045 & 72.666 & 98.648 \\
\midrule
\ding{51} & \ding{51} & \ding{51} & \textbf{32.205} & \textbf{72.734} & \textbf{98.685} \\
\bottomrule
\end{tabular}
\end{table}

\noindent\textbf{Motion estimation consistency.} For each endpoint pair, we compare the backward displacement used to warp $I_0$ to the midpoint across training runs, averaging pairwise Euclidean differences over voxels and cases. This evaluates reproducibility of the recovered correspondence for the same observations, complementing tests of reconstructed appearance. Our method has low disagreement, comparable to MPVF and substantially below Fourier-Net+ and DDM in Table~\ref{tab:cross_run_consistency}. Endpoint image agreement can conceal different underlying correspondences. Small field differences together with accurate reconstruction support a stable choice of motion explanation, consistent with the intended restriction of motion ambiguity.

\begin{table}[!t]
\centering
\caption{Mean pairwise displacement difference across training runs on ACDC at $t=0.5$ ($\mathrm{voxel}$, $\downarrow$).}
\label{tab:cross_run_consistency}
\small
\setlength{\tabcolsep}{3pt}
\renewcommand{\arraystretch}{1.1}
\begin{tabular}{@{}lcccc@{}}
\toprule
 & Ours & MPVF & Fourier-Net+ & DDM \\
\midrule
Difference & 0.03989 & 0.04572 & 0.14694 & 0.23164 \\
\bottomrule
\end{tabular}
\end{table}

\begin{figure}[!tp]
\centering
\includegraphics[width=\columnwidth]{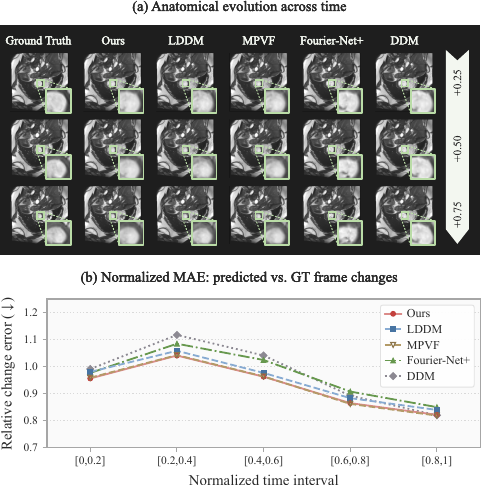}
\caption{Temporal reconstruction on ACDC. (a) Ventricular anatomy and enlarged details across intermediate query times. (b) Relative change error across normalized time intervals, evaluating agreement with the reference sequence's image evolution. Lower values indicate more faithful temporal reconstruction.}
\label{fig:temporal_analysis}
\end{figure}

\begin{figure}[!t]
\centering
\includegraphics[width=\columnwidth]{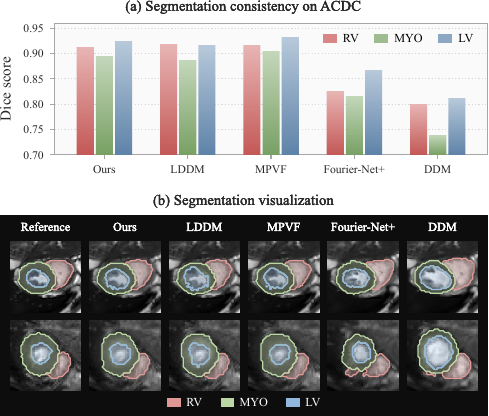}
\caption{Segmentation consistency on ACDC. (a) Dice agreement between masks predicted from reconstructed and reference volumes, averaged over test volumes including endpoints. RV, MYO, and LV denote the right ventricular cavity, left ventricular myocardium, and left ventricular cavity. (b) Reference and reconstructed images with their model-predicted masks.}
\label{fig:segmentation_consistency}
\end{figure}

\noindent\textbf{Temporal reconstruction fidelity.} As shown in Fig.~\ref{fig:temporal_analysis}(a), our reconstructions follow the evolving ventricular contours and blood-pool appearance, while local boundaries and internal contrast are less faithfully recovered by LDDM and Fourier-Net+. This comparison tests whether detail preserved at one query also changes appropriately at other times, including the relative positions of local boundaries as the ventricle deforms. A shared velocity representation links these reconstructions, allowing the observed anatomy to evolve coherently across intermediate queries.

Relative change error measures the mean absolute discrepancy between predicted and reference adjacent-volume change rates, normalized by each case's time-averaged absolute reference-change rate. Rates use the actual time separation between volumes, with spatial averaging over anatomy identified by intensity variation in the reference sequence. Our method closely tracks MPVF and remains below LDDM and Fourier-Net+ across the intervals in Fig.~\ref{fig:temporal_analysis}(b), and below DDM in four of five intervals. These results complement frame-wise similarity: the preserved ventricular detail is accompanied by competitive agreement with reference change rates across the sampled intervals, linking anatomical appearance to the progression of image change over time.

\noindent\textbf{Segmentation consistency.} A standard 3D U-Net~\cite{cicek2016three} trained on ACDC training cases is frozen and applied to both reconstructed and reference volumes; Dice measures agreement between its predicted masks. Our method approaches MPVF and exceeds Fourier-Net+ and DDM across the three structures in Fig.~\ref{fig:segmentation_consistency}(a). Its advantage over LDDM in MYO and LV indicates that maintaining blood-pool appearance alone does not ensure consistent myocardial extent. Cavity and myocardial masks depend on neighboring boundaries, so a local contour shift can alter their relative extent even when the overall appearance changes little. Agreement across these regions supports preservation of their relative configuration during interpolation. The overlays in Fig.~\ref{fig:segmentation_consistency}(b) connect reconstruction fidelity with consistent structural delineation.

\noindent\textbf{Computational cost.} Our method provides higher query throughput than MPVF, DDM, and LDDM, with estimated training time close to DDM, as shown in Table~\ref{tab:efficiency_profiles}. Training uses the dense head to build motion evidence; query-time synthesis uses the shared reference and predicted motion coordinates without this auxiliary head. Fourier-Net+ is faster, but the reconstruction comparisons favor our method. The resulting trade-off places the framework between fast band-limited synthesis and more expensive motion-guided or generative alternatives, retaining sample-specific motion expression at a moderate synthesis cost and making that flexibility practical for repeated queries in temporally dense reconstruction.

\begin{table}[!t]
\centering
\caption{Computational cost on ACDC. QPS counts complete 3D midpoint queries. MPVF training time includes all training stages.}
\label{tab:efficiency_profiles}
\small
\setlength{\tabcolsep}{3pt}
\renewcommand{\arraystretch}{1.1}
\begin{tabular}{@{}lD{.}{.}{2.3}D{.}{.}{3.2}D{.}{.}{3.2}@{}}
\toprule
Method & \multicolumn{1}{c}{\#P (M)} & \multicolumn{1}{c}{\shortstack{Est. train time\\(s/epoch) $\downarrow$}} & \multicolumn{1}{c}{QPS $\uparrow$} \\
\midrule
Ours & 15.100 & 19.18 & 40.52 \\
MPVF & 14.536 & 105.14 & 4.61 \\
DDM & 23.164 & 18.42 & 14.04 \\
LDDM & 15.653 & 51.52 & 2.72 \\
Fourier-Net+ & 12.414 & 1.84 & 121.16 \\
\bottomrule
\end{tabular}
\end{table}

\section{Conclusion}

We presented a framework that learns a motion manifold to constrain ambiguity in unsupervised 4D medical image interpolation. Cross-sample spectral basis learning captures recurring motion structure from endpoint-derived estimates, while sample-conditioned Cayley reorientation varies the effective motion directions without increasing the number of independent motion coordinates. The resulting representation accommodates sample-specific variation within a restricted motion family. Experiments on ACDC and 4D-Lung demonstrate state-of-the-art performance among the evaluated methods trained without intermediate volumes and competitive results against methods using intermediate supervision. Module analyses, temporal evaluation, and motion-estimation consistency link these reconstruction results to effective use of the shared directions and reproducible motion estimates. Learning a shared motion family while adapting its expression to individual observations provides a useful representation-level constraint on ambiguity in endpoint-supervised interpolation.

\bibliographystyle{IEEEtran}
\bibliography{refer}

\end{document}